\RequirePackage{fix-cm}
\documentclass[smallextended]{svjour3}       
\smartqed  

\usepackage{graphicx}
\usepackage{amsmath}
\usepackage{amssymb}
\usepackage{algorithmic}
\usepackage{algorithm}
\usepackage{subfigure}
\usepackage{array}
\usepackage{tabularx}
\usepackage{multirow}
\usepackage{pdfsync}
\usepackage[pagewise]{lineno}

\newcommand{\bfx}{{\textbf{x}}}

\newcommand{\bfw}{{\textbf{w}}}
\newcommand{\bfy}{{\textbf{y}}}

\begin{document}

\title{Manifold regularization in structured output space for semi-supervised structured output prediction}

\author{Fei Jiang \and Lili Jia \and Xiaobao Sheng \and Riley LeMieux}

\institute{F. Jiang\at
College of Fine Arts, Shanghai University, Shanghai 200444, China
\and
X. Sheng\at
School of Economics and Management, Tongji University, Shanghai 200092, China\\
\email{sxbao765@hotmail.com}\\
Corresponding author
\and
L. Jia, X. Sheng
\at
The 3rd Research Institute of Ministry of Public Security, Shanghai 200031, China
\and
R. LeMieux\at
Department of Computing and Information Sciences, Kansas State University, Manhattan, KS 66506, United States\\
\email{riley.lemieux@hotmail.com}
}

\date{Received: date / Accepted: date}

\maketitle

\begin{abstract}
Structured output prediction aims to learn a predictor to predict a structured output from a input data vector. The structured outputs include vector, tree, sequence, etc. We usually assume that we have a training set of input-output pairs to train the predictor. However, in many real-world applications, it is difficult to obtain the output for a input, thus for many training input data points, the structured outputs are missing. In this paper, we discuss how to learn from a training set composed of some input-output pairs, and some input data points without outputs. This problem is called semi-supervised structured output prediction. We propose a novel method for this problem by constructing a nearest neighbor graph from the input space to present the manifold structure, and using it to regularize the structured output space  directly.  We define a slack structured output for each training data point, and proposed to predict it by learning a structured output predictor. The learning of both slack structured outputs and the predictor are unified within one single minimization problem. In this problem, we propose to minimize the structured loss between the slack structured outputs of neighboring data points, and the prediction error measured by the structured loss. The problem is optimized by an iterative algorithm. Experiment results over three benchmark data sets show its advantage.
\end{abstract}

\section{Introduction}

\subsection{Background}

In machine learning community, the problems of pattern classification and regression has been studied well. Classification and regression are two most popular supervised learning problems \cite{xia2015study,Oonk201580,Braida20154733,wang2014effective,zhang2011empirical,luo2011based,wang2010data,wang2012mathematical,web14,cross13,wang2014next,wang2014optimizing,wang2015towards,wang2015representing,wang2015multiple}. In these problems, we usually have a training set of input-output pairs. The task is to train a predictive model from the training set to predict the output of a test input. In both the problems of classification and regression, the input is usually a feature vector. The output of classification problems is a binary class label, which represents a positive class or a negative class. The output of regression problems is a continues response variable. Recently, it is proposed that the output of a machine learning problem can be beyond a binary label and a continues response, and the output is structured in many real-world applications \cite{Astikainen2011367,Han20141665,Su201038,Kajdanowicz2011333,Joachims20061,Wu20092087,Kim2010649}. For example, in multi-class classification problems, the output is a vector presenting which class the input belongs to. In hierarchical classification problems, the classes are organized as a tree, and each class is a node of the tree. Moreover, in natural language parsing problems, the output of a input language sequence is a sequence. When the structured output is considered, the transitional predictive model learning algorithms cannot be used because the  output does not to them. To solve this problem, the structured output prediction problem is proposed to learn a specific given structured output. This problem assume a training set of input-structured output pairs are available for the learning problem. However, in real-world applications, it is usually expensive or time-consuming to obtain a structured output for a input data point. Thus in many cases, we have a limited number of input-structure output pairs, and a large number of inputs without corresponding structured outputs. In this case, we try to learn a predictive model with a large number of input data points and a small number of structured outputs. This problem is call semi-supervised structured output prediction \cite{Brefeld2006145a,Suzuki2007791,Li20143205}. In this paper, we investigate this problem, and proposed a novel method to solve it.

\subsection{Related works}

There are some existing works on semi-supervised structured output prediction problem. We introduce them as follows.

\begin{itemize}
\item Altun et al. \cite{Altun200533}  proposed the problem of predicting multiple inter-dependent outputs by learning in a semi-supervise setting, and a method to solve this problem. The method is a maximum-margin method, and it uses the manifold of the input data space by exploring both the labeled and unlabeled data points. Moreover, this method is a inductive method and it learns a predictive model to predict the structured outputs for new coming test data points.

\item Brefeld and Scheffer \cite{Brefeld2006145a} proposed a method for semi-supervised learning for structured output prediction. The method is a co-training method, and it is based on learning in a joint input-output space. It maximizes the consensus among different independent hypotheses, and extend it to a semi-supervised support vector machine learning algorithm in the joint input-output space. Moreover, the prediction loss of structured output is measured by a arbitrary structured loss function.

\item Suzuki et al.  \cite{Suzuki2007791} proposed a semi-supervised structured output prediction method for sequence labeling task. This method is based on a combination of both generative and discriminative models. The objective of this method is constructed as a log-linear form, and the objective is a combination of discriminative structured predictor and generative model to use the input data points without structured output (unlabeled data points).  Moreover, these unlabeled data points is utilized by the  generative model to increase the sum of the discriminant functions for all outputs.

\item Li and Zemel \cite{Li20143205} proposed a max-margin method for semi-supervised structured output prediction problem. This methods can use the both the discrete optimization algorithms and high order regularization based on the unlabeled data points. This method is shown to be closely relevant to the Posterior Regularization.
\end{itemize}

Manifold learning is a popular topic in semi-supervised learning problems \cite{Ho2015,Xing2015395,Feng2015,Lorente201517}. It impose that if two data points are neighboring in the input space, their outputs should also be close to each other. Because the outputs of the data points are not complete, and most of the outputs of training data points are missing, it is important to infer the missing output from the available outputs by using the neighborhood relationship in the input space. Manifold learning has been a powerful regularization method in both classification and regression problems, and usually a squared $\ell_2$ norm distance is used to measure how close two outputs (binary labels, or continues responses) are. However, in structured output prediction problem, the squared $\ell_2$ norm distance cannot fit the structured outputs. In \cite{Altun200533}, a manifold regularization is also used. However, due to the complexity of the structured outputs, the regularization is not performed directly in the output space, but to the ``parts'' of the joint input-output space. A pair of ``parts'' is also compared using the squared $\ell_2$ norm, so that the regularization term will not bring difficulty to the optimization of the problem. It is not guaranteed that regularizing the ``parts''  of input-output space can lead to the neighborhood smoothness in the output space. Actually, we can measure how close a pair of structured outputs are by a predefined structured loss function. However, due to the complexity of this loss function, it is very difficult to optimize it to solve the parameter of the predictor.

\subsection{Our contributions}

To solve the problem mentioned above, in this paper, we propose to regularize the structured outputs directly in the structured output space. To avoid the difficulty of optimizing the structured loss function, we introduce a slack structured output for each training data point. This slack structured output presents the optimal output, and it is also treated as a variable during the learning procedure. For the labeled data points, their true structured outputs are available, we impose their slack structured outputs to be consistent with their true structured outputs. To prorogate the structured output from the labeled data to the unlabeled data, we use the manifold information to present the connections between the data points. To present the manifold information, we construct a nearest neighbor graph in the input data space, and use it to regularize the output space directly. More specifically, if the inputs of two data points are neighbors, we also hope their slack structured outputs are close to each other. We use the structured loss function to measure how the compared structured outputs are close to each other. Moreover, to learn the predictive model, we learn the model parameter to fit the model to the slack structured outputs. In this way, we impose the slack structured outputs to be consistent to both the prediction results of the predictive model, and the structured outputs of its nearest neighbors.

The predictive model is designed as a linear function of a joint input-output representation. We construct a objective function with respect to both the slack structured outputs and the predictive model parameter. In this objective function, we minimize the losses of the prediction results of the predictive model against the slack structured outputs, and the losses of the structured outputs of each pair of neighboring data points, simultaneously. The objective is optimized by an iterative algorithm, and the slack structured outputs and the predictive model parameter are updated alternately.

The contributions of this work are of two folds:

\begin{enumerate}
\item We solve the problem of manifold regularization in structured output space by introducing a slack structured output for each data point, both labeled and unlabeled, and comparing a pair of structured outputs of neighboring data points  by the structured loss function.

\item We propose a novel iterative algorithm to solve the slack structured outputs and the predictive model parameters simultaneously. The optimization of the slack structured outputs are regularized by both the predictive model and the manifold. Moreover, we develop an efficient gradient descent-based method to update the predictive model parameter. This method is more efficient than the most popular optimization algorithm used in structured output prediction methods, cutting plane algorithm \cite{Chouman201599,Fang2015212,Eronen2015641,Abdelouadoud20159}, because it avoids the time-consuming quadratic programming problem of cutting plane algorithm.
\end{enumerate}

\subsection{Paper organization}

The rest parts of this paper are organized as follows. In section \ref{sec:method} we introduce the proposed method, by first modeling the problem as a minimization problem, then solving it using an alternate optimization strategy, and finally developing an iterative algorithm. In section \ref{sec:experiments}, the proposed is studied experimentally. It is compared to state-of-the-art semi-supervised structured output prediction methods. Its sensitivity to parameter and running time is also studied. In section \ref{sec:conclusion}, we give the conclusion and the future works.

\section{Proposed method}
\label{sec:method}

In this section, we introduce the proposed method. The problem is modeled as a formulation of minimization problem, and it is then solved by a alternate optimization method with an iterative algorithm.

\subsection{Problem modeling}

We consider a problem of structured output prediction problem, where the input is a  $d$-dimensional input vector, $\bfx\in \mathbb{R}^d$, and the output is a structured output, $y\in \mathcal{Y}$, where $\mathcal{Y}$ is the structured output space. We assume we have a training set of data points $\mathcal{X} = \{\theta_i\}_{i=1}^n$, where $\theta_i$ is the $i$-th data point, and $n$ is the number of the data points in $\mathcal{X}$. $\mathcal{X}$ is composed of two subsets, $\mathcal{X}=\mathcal{L}\bigcup \mathcal{U}$, where $\mathcal{L}$ is the labeled data point set, and $\mathcal{U}$ is the unlabeled data point set. The data points of $\mathcal{L}$ are presented as a input-output pairs, $\theta_i=(\bfx_i,y_i)|_{i:\theta_i\in \mathcal{L}}$, where $\bfx_i\in \mathbb{R}^d$  input vector of the $i$-th data point, $y_i\in \mathcal{Y}$ is its corresponding structured output. The data points in $\mathcal{U}$ only have the inputs while the structured outputs are missing, $\theta_i=\bfx_i|_{i:\theta_i\in \mathcal{U}}$.
To learn the missing structured outputs for the data points in $\mathcal{U}$, and a predictive model to predict the structured output for a test input, we consider the following problems to model the objective function.

\subsubsection{Regularizing the structured outputs by manifold}

We want to regularize the structured output by the manifold, but for the data points in $\mathcal{U}$, the structured outputs are missing. To solve this problem, we introduce a slack structured output, $z_i\in \mathcal{Y}$, for each data point $\theta_i|_{i:\theta_i \in \mathcal{X}}$. This slack structured output $z_i$ presents the optimal output we want to learn for the $i$-th data point.

For a labeled data point, $\theta_i|_{i:\theta_i\in \mathcal{L}}$, since its true structured output $y_i$ is known, we impost $z_i = y_i$. For these unlabeled data points, $\theta_i|_{i:\theta_i\in \mathcal{U}}$, we want to predict their slack outputs by prorogating the output information from the labeled data points via a manifold. To present the manifold information, we construct a nearest neighbor graph from the input of data points of $\mathcal{X}$. For the input vector $\bfx_i$ of each data point $\theta_i$ from the inputs of data points in $\mathcal{X}$, we find its $K$ nearest neighbors and denote the set of its nearest neighbors as $\mathcal{N}_i$. To construct the graph, we treat each data point as a node of the graph, and put a edge between the $i$-th node and the $j$-th node if $\bfx_j\in \mathcal{N}_i$. Denoting $\mathcal{E}$ as the set of edges, we have

\begin{equation}
\label{equ:E}
\begin{aligned}
\mathcal{E}=\{(\theta_i,\theta_j):\theta_i,\theta_j\in \mathcal{X}, \bfx_j\in \mathcal{N}_i\}.
\end{aligned}
\end{equation}
The weight of the egde $(\theta_i,\theta_j)$, $\omega_{ij}$, is assigned as a Gaussian kernel of the Euclidian distance between $\bfx_i$  and $\bfx_j$,

\begin{equation}
\label{equ:W}
\begin{aligned}
\omega_{ij}=
\left\{\begin{matrix}
\exp\left (- \frac{\|\bfx_i-\bfx_j\|_2^2}{2\sigma}\right ), &if~(\theta_i,\theta_j)\in \mathcal{E} \\
0, &otherwise.
\end{matrix}\right.
\end{aligned}
\end{equation}
$\omega_{ij}$ is a measurement of the similarity between a pair of neighboring data points in the input space. We try to map the similarity relationship from the input space to the structured space. For a pair neighboring data points, if there are similar in the input space, i.e., $\omega_{ij}$ is large, their structured outputs should also be similar to each other, i.e., $z_i$ and $z_j$ are close to each other. To measure how $z_i$ and $z_j$ are close to each other, we use a structured loss function, $\Delta(z_i,z_j)$, to compare $z_i$ against $z_j$. $\Delta(z_i,z_j)$ is a loss function to measure the loss if a structured label $z_j$ is wrongly predicted as $z_i$. For example, when the structured output are the nodes of a tree, $\Delta(z_i,z_j)$ is defined as the height of the first common ancestor of $z_i$ and $z_j$ in the tree.  Naturally, if $\omega_{ij}$ is large, we hope $\Delta(z_i,z_j)$ is as mall as possible.  Thus we propose to minimize $\Delta(z_i,z_j)$ weighted by $\omega_{ij}$ with regard to $z_i$ and $z_j$,

\begin{equation}
\label{equ:reg}
\begin{aligned}
\min_{z_1,\cdots,z_n}
~&
\left \{M(z_1,\cdots,z_n)=
\sum_{i,j:(\theta_i,\theta_j)\in \mathcal{E}} \omega_{ij} \Delta(z_i,z_j)
\right \},\\
s.t.~&
z_i=y_i, \forall ~i:\theta_i\in \mathcal{L}.
\end{aligned}
\end{equation}
In this way, we regularize the learning of slack structured outputs directly by the manifold, instead of regularizing the joints input-output space.

\subsubsection{Learning predictive model}

The problem of structured output prediction is to learn a predictive model $f$ to predict a true structured output $y\in \mathcal{Y}$ from a input $\bfw\in \mathbb{R}^d$,

\begin{equation}
\label{equ:predictive}
\begin{aligned}
y\leftarrow f(\bfx;\bfw)
\end{aligned}
\end{equation}
where $\bfw$ is parameter of the predictive model $f$. To design the predictive model, we present a joint representation function to match a input $\bfx$ against a candidate structured output $y'\in \mathcal{Y}$, $\Phi(\bfx,y') \in \mathbb{R}^{m}$, where $m$ is the dimension of the joint representation. An example of this representation function is for the vector output, where $y'$ is a vector, and $\Phi(\bfx,y')=\bfx\bigotimes y'$ is the Hadamard product of $\bfx$ and $y'$. We further design a matching function, $g(\bfx,y';\bfw)$,  to obtain the matching score of $\bfx$ and $y'$,

\begin{equation}
\label{equ:g}
\begin{aligned}
g(\bfx,y';\bfw) = \bfw^\top \Phi(\bfx,y')
\end{aligned}
\end{equation}
where $\bfw\in \mathbb{R}^m$ is the parameter of the matching function. The predictive model is based on the matching function, and it returns the optimal candidate structured output, $y^*$, that maximized the matching scores,

\begin{equation}
\label{equ:g}
\begin{aligned}
y^* \leftarrow f(\bfx;\bfw)= {\arg\max}_{y'\in \mathcal{Y}} \bfw^\top \Phi(\bfx,y')
\end{aligned}
\end{equation}
The prediction error can can be measured by a loss function, $\Delta(y^*,y)$, to compare the predicted structured output, $y^*$, against the true structured output, $y$. The problem of structured output prediction is changed to the learning of the parameter vector $\bfw$.

Since for the data points in $\mathcal{U}$, the true structured outputs are missing, we use the slack structured outputs to guide the learning of the model parameter. We hope with the learned parameter vector, $\bfw$, for the $i$-th training data point, the loss of predicting $z_i$ as $y_i^*$, $\Delta(y_i^*, z_i)$, can be minimized. Thus we have the following optimization problem,

\begin{equation}
\label{equ:loss1}
\begin{aligned}
\min_{\bfw,z_1,\cdots,z_n}
~&
 \sum_{i=1}^n \Delta(y_i^*,z_i)
,\\
s.t.~&
z_i=y_i, \forall ~i:\theta_i\in \mathcal{L}.
\end{aligned}
\end{equation}
where $y_i^*$ is the predicted structured output of the $i$-th data point.

Due to the complexity of the loss function $\Delta$, this problem is hard to optimize with regard to $\bfw$ directly. Instead of minimizing $\Delta(y_i^*,z_i)$ directly, we seek and minimize its upper bound. According to (\ref{equ:g}),

\begin{equation}
\label{equ:upper}
\begin{aligned}
&\bfw^\top \Phi(\bfx_i,y_i^*) \geq \bfw^\top \Phi(\bfx_i,z_i), \forall~ z_i\in \mathcal{Y},\\
&\Rightarrow \bfw^\top \left ( \Phi(\bfx_i,y_i^*) -\Phi(\bfx_i,z_i)\right )+ \Delta(y_i^*,z_i) \geq \Delta(y_i^*,z_i).
\end{aligned}
\end{equation}
We replace the predicted structured output $y_i^*$ in (\ref{equ:upper}) by a strutted output $y_i''$ to maximize the left hand of the list line of (\ref{equ:upper}), so that

\begin{equation}
\label{equ:upper1}
\begin{aligned}
&\max_{y_i''\in \mathcal{Y}}
\left [
\bfw^\top \left ( \Phi(\bfx_i,y_i'') -\Phi(\bfx_i,z_i)\right )+ \Delta(y_i'',z_i)
 \right ] \\
& \geq \bfw^\top \left ( \Phi(\bfx_i,y_i^*) -\Phi(\bfx_i,z_i)\right )+ \Delta(y_i^*,z_i)\\
& \geq \Delta(y_i^*,z_i).
\end{aligned}
\end{equation}
Thus a upper bound of $\Delta(y_i^*,z_i)$ is obtained as follows,

\begin{equation}
\label{equ:upper2}
\begin{aligned}
&\max_{y_i''\in \mathcal{Y}}
\left [
\bfw^\top \left ( \Phi(\bfx_i,y_i'') -\Phi(\bfx_i,z_i)\right )+ \Delta(y_i'',z_i)
 \right ] \\
 &=
\bfw^\top \left ( \Phi(\bfx_i,\upsilon_i) -\Phi(\bfx_i,z_i)\right )+ \Delta(\upsilon_i,z_i),
\end{aligned}
\end{equation}
where $\upsilon_i$ is the structured output that maximize the left hand of (\ref{equ:upper2}),

\begin{equation}
\label{equ:upper3}
\begin{aligned}
&\upsilon_i = {\arg\max}_{y_i''\in \mathcal{Y}}
\left [
\bfw^\top \left ( \Phi(\bfx_i,y_i'') -\Phi(\bfx_i,z_i)\right )+ \Delta(y_i'',z_i)
 \right ].
\end{aligned}
\end{equation}

Replacing $\Delta(y_i^*,z_i)$ by its upper bound in (\ref{equ:upper2}), we rewrite (\ref{equ:loss1}) as

\begin{equation}
\label{equ:loss2}
\begin{aligned}
\min_{\bfw,z_1,\cdots,z_n}
~&\left \{
L(\bfw,z_1,\cdots,z_n) =
 \sum_{i=1}^n \left [
 \bfw^\top \left ( \Phi(\bfx_i,\upsilon_i) -\Phi(\bfx_i,z_i)\right )+ \Delta(\upsilon_i,z_i)
 \right ]
\right \},\\
s.t.~&
z_i=y_i, \forall ~i:\theta_i\in \mathcal{L}.
\end{aligned}
\end{equation}
In this way, we transfer the problem of minimizing $\Delta(y_i^*,z_i)$ to the minimization of its upper bound.

\subsubsection{Reducing the model complexity}

To avoid the over-fitting problem, we try to reduce the complexity of the model. The complexity of the model can be measured by the squared $\ell_2$ norm of the model parameter vector, $\|\bfw\|_2^2$. To reduce the complexity, we propose to minimize a regularization term $R(\bfw)$,

\begin{equation}
\label{equ:w}
\begin{aligned}
\min_\bfw \left \{ R(\bfw) = \frac{1}{2} \|\bfw\|_2^2 \right \}.
\end{aligned}
\end{equation}

\subsubsection{Overall optimization problem}

The overall optimization problem of the proposed method is a combination of the three terms in (\ref{equ:reg}), (\ref{equ:loss2}), and (\ref{equ:w}),

\begin{equation}
\label{equ:object}
\begin{aligned}
\min_{\bfw, z_1,\cdots,z_n}
~&
\left \{
O(\bfw,z_1,\cdots,z_n)
\vphantom{\frac{1}{1}}
\right .
\\
&
= M(z_1,\cdots,z_n)+C_1 L(\bfw,z_1,\cdots,z_n)  +C_2 R(\bfw)
\\
&=
\sum_{i,j:(\theta_i,\theta_j)\in \mathcal{E}} \omega_{ij} \Delta(z_i,z_j)\\
&
\left .
+C_1  \sum_{i=1}^n \left [
 \bfw^\top \left ( \Phi(\bfx_i,\upsilon_i) -\Phi(\bfx_i,z_i)\right )+ \Delta(\upsilon_i,z_i)
 \right ]
+ \frac{C_2}{2} \|\bfw\|_2^2
\right \},\\
s.t.~&
z_i=y_i, \forall ~i:\theta_i\in \mathcal{L},
\end{aligned}
\end{equation}
where $C_1$ and $C_2$ are the tradeoff parameters. The first term of the objective function is to regularize the slack structured outputs by the manifold, the second term is to reduce the loss of prediction error, and the last term is to reduce the complexity of the model. In this problem, the learning of the slack structured outputs are regularized by three information sources: the manifold, the known true structured outputs of the labeled data points, and the prediction results of the predictive model.

\subsection{Problem optimization}

To solve the problem in (\ref{equ:object}), we use an alternate optimization strategy. In an iterative algorithm, when the model parameter vector $\bfw$ is considered, the slack structured outputs $z_1,\cdots,z_n$ are fixed. When $z_1,\cdots, z_n$ are considered, $\bfw$ is considered. In the following subsections, we will discuss how to solve $\bfw$ and $z_1,\cdots, z_n$ respectively.

\subsubsection{Solving $\bfw$ while fixing $z_1,\cdots,z_n$}

When we consider the model parameter vector $\bfw$, the slack structured outputs $z_1,\cdots, z_n$ are fixing. We remove the terms in (\ref{equ:object}) irrelevant to $\bfw$, and obtain the following problem,

\begin{equation}
\label{equ:object1}
\begin{aligned}
\min_{\bfw}
~&
\left \{
O_1(\bfw)
\vphantom{\frac{1}{1}}
=
C_1  \sum_{i=1}^n \left [
 \bfw^\top \left ( \Phi(\bfx_i,\upsilon_i) -\Phi(\bfx_i,z_i)\right ) + \Delta(\upsilon_i,z_i)
 \right ]
+ \frac{C_2}{2} \|\bfw\|_2^2
\right \}.
\end{aligned}
\end{equation}
Please note that $\upsilon_i$ is also a function of $\bfw$ according to (\ref{equ:upper3}). However, because it is coupled with a maximization problem, thus it is hard to optimize with regard to $\bfw$ directly. Thus we use the strategy of expectation-maximization algorithm, update $\upsilon_i$ by using the solutions of $\bfw$ and $z_1,\cdots, z_n$ in previous iteration, and then fix it when $\bfw$ is optimized in current iteration. After $\upsilon_i$ is fixed, we use the gradient descent algorithm to update $\bfw$. To seek the minimization of $O_1(\bfw)$, $\bfw$ should descent to the direction of gradient. The gradient function of $O_1(\bfw)$ is

\begin{equation}
\label{equ:gradient}
\begin{aligned}
\nabla O_1(\bfw) =
C_1  \sum_{i=1}^n
\left ( \Phi(\bfx_i,\upsilon_i) -\Phi(\bfx_i,z_i)\right )
+ C_2 \bfw.
\end{aligned}
\end{equation}
The updating rule is

\begin{equation}
\label{equ:update}
\begin{aligned}
\bfw
\leftarrow &\bfw - \eta \nabla O_1(\bfw)\\
 =&\bfw - \eta \left [
C_1  \sum_{i=1}^n
\left ( \Phi(\bfx_i,\upsilon_i) -\Phi(\bfx_i,z_i)\right )
+ C_2 \bfw
\right ]\\
=&(1-\eta C_2) \bfw - \eta C_1
\sum_{i=1}^n
\left ( \Phi(\bfx_i,\upsilon_i) -\Phi(\bfx_i,z_i)\right ),
\end{aligned}
\end{equation}
where $\eta$ is the descent step.

\subsubsection{Solving $z_1,\cdots,z_n$ while fixing $\bfw$}

We fix the $\bfw$ when $z_1,\cdots,z_n$ are considered, and remove the terms irrelevant to $z_1,\cdots,z_n$. The following problem is obtained,

\begin{equation}
\label{equ:object3}
\begin{aligned}
\min_{z_1,\cdots,z_n}
~&
\left \{
O_2(z_1,\cdots,z_n)
=
\sum_{i,j:(\theta_i,\theta_j)\in \mathcal{E}} \omega_{ij} \Delta(z_i,z_j)
\right .
\\
&
\left .
+C_1  \sum_{i=1}^n \left [
- \bfw^\top \Phi(\bfx_i,z_i)+ \Delta(\upsilon_i,z_i)
 \right ]
\right \},\\
s.t.~&
z_i=y_i, \forall ~i:\theta_i\in \mathcal{L}.
\end{aligned}
\end{equation}
It is difficult to optimize all the slack structured outputs $z_1, \cdots, z_n$ simultaneously. Thus we chose to update them one by one. When one slack structured output $z_i$ is considered, other ones $z_j|_{j\neq i}$ are fixed. In this case, we obtain the following problem for the $i$-th data point,

\begin{equation}
\label{equ:object4}
\begin{aligned}
\min_{z_i}
~&
\left \{
O_3(z_i)
=
\sum_{j:(\theta_i,\theta_j)\in \mathcal{E}} \omega_{ij} \Delta(z_i,z_j)
+
\sum_{j':(\theta_{j'},\theta_i)\in \mathcal{E}} \omega_{j'i} \Delta(z_{j'},z_i)
\right .
\\
&\left.
+C_1  \left [
- \bfw^\top \Phi(\bfx_i,z_i)+ \Delta(\upsilon_i,z_i)
 \right ]
 \vphantom{\sum_1^2}
\right \},\\
s.t.~&
z_i=y_i, \forall ~i:\theta_i\in \mathcal{L}.
\end{aligned}
\end{equation}
From the formulation, we can see that the optimal $z_i$ should be consistent to the slack structured outputs of its nearest neighbors, and the prediction result of the predictive model. The solution for this problem can be obtained by a linear search in the structured output space,

\begin{equation}
\label{equ:solution}
\begin{aligned}
z_i =
\left\{\begin{matrix}
{\arg\max}_{y'\in \mathcal{Y}} O_3(y'_i),&if~\theta_i\in \mathcal{U} \\
y_i ,& otherwise.
\end{matrix}\right.
\end{aligned}
\end{equation}

\subsection{Iterative algorithm}

We summarize the developed iterative learning algorithm in Algorithm (\ref{alg:iter}). From this algorithm, we can see that the iterations are repeated $T$ times. In each iteration, we first update $\upsilon_i$ and $z_i$ for each data point, and then update $\bfw$. This algorithm is named as manifold regularized structured output learning algorithm (MRSO).

\begin{algorithm}[h!]
\caption{Iterative algorithm of MRSO.}
\label{alg:iter}
\begin{algorithmic}
\STATE \textbf{Input}: Training set of data points $\mathcal{X}$;
\STATE \textbf{Input}: Tradeoff parameters $C_1$ and $C_2$;
\STATE \textbf{Input}: Maximum number of iterations, $T$;

\STATE Initialize model parameter vector $\bfw^0$;

\STATE Initialize the slack structured outputs $z_1^0,\cdots,z_n^0$;

\FOR{$t=1,\cdots,T$}

\FOR{$i=1,\cdots,n$}

\STATE Update $\upsilon_i^t$ of the $i$-th data point by fixing $z_i^{t-1}$ and $\bfw^{t-1}$,

\begin{equation}
\label{equ:upsilon}
\begin{aligned}
&\upsilon_i^t = {\arg\max}_{y_i''\in \mathcal{Y}}
\left [
{\bfw^{t-1}}^\top \left ( \Phi(\bfx_i,y_i'') -\Phi(\bfx_i,z_i^{t-1})\right )+ \Delta(y_i'',z_i^{t-1})
 \right ].
\end{aligned}
\end{equation}

\STATE Update $z_i^t$ of the $i$-th data point by fixing $\bfw^{t-1}$, $z_{j}^{t-1}|_{j\neq i}$ and $\upsilon_i^t$.

\IF{$\theta_i \in \mathcal{U}$}

\STATE

\begin{equation}
\label{equ:z}
\begin{aligned}
z_i^t =
& {\arg\min}_{y_i'\in \mathcal{Y}}
~
\left \{
\sum_{j:(\theta_i,\theta_j)\in \mathcal{E}} \omega_{ij} \Delta(y_i',z_j^{t-1})
+
\sum_{j':(\theta_{j'},\theta_i)\in \mathcal{E}} \omega_{j'i} \Delta(z_{j'}^{t-1},y_i')
\right.
\\
&
\left .
+C_1  \left [
- {\bfw^{t-1}}^\top \Phi(\bfx_i,y_i')+ \Delta(\upsilon_i^{t},y_i')
 \right ]
 \vphantom{\sum_{1_1}^2}
 \right \};
\end{aligned}
\end{equation}

\ELSE

\STATE $z^t_i=y_i$;

\ENDIF

\ENDFOR

\STATE Update $\bfw^t$ by fixing $\upsilon_1^t,\cdots,\upsilon_n^t$, and $z_1^t,\cdots,z_n^t$,

\begin{equation}
\label{equ:w}
\begin{aligned}
\bfw^t = (1-\eta C_2) \bfw^{t-1} - \eta C_1
\sum_{i=1}^n
\left ( \Phi(\bfx_i,\upsilon_i^t) -\Phi(\bfx_i,z_i^t)\right );
\end{aligned}
\end{equation}

\ENDFOR

\STATE \textbf{Output}:  $\bfw^T$ and $z_1^T,\cdots,z_n^T$.

\end{algorithmic}
\end{algorithm}

\section{Experiments}
\label{sec:experiments}

\subsection{Data sets}

\begin{itemize}
\item The first data set we used is Cora data set \cite{senaimag08}. The output of this data set is the class label vector of multi-class classification problem. This data set is a linked computer science paper data set. Each paper is a treated as a data point. In this data set, there are 9,947 data points. The papers without a reference list is removed from the data set, and 9,555 papers are left. All the papers belong to the 8 classes. To construct a feature vector from a paper, we extract a term frequency vector, and a link view vector, and concatenate them as a feature vector \cite{zhang2012automated}.

    For each data point, $\bfx_i$, we construct a vector output, $\bfy_i=[y_{i1},\cdots,y_{i8}]\in \{1,0\}^8$, as the structured output. This vector, $\bfy_i$, is a $8$-dimensional binary vector. If this data point belongs to the $k$-class, then the $k$-th element of this vector is 1, or 0 otherwise,

\begin{equation}
\label{equ:y_i}
\begin{aligned}
y_{ik}=
\left\{\begin{matrix}
1, &if~\bfx_i~belongs~to~the~k-th~class, \\
0, &otherwise.
\end{matrix}\right.
\end{aligned}
\end{equation}
We further define the joint input-output representation function as $\Phi(\bfx,\bfy) = \bfx\otimes \bfy$. To measure the prediction error of predicting $\bfy_i$ as $\bfy_i^*$ by the $0-1$ loss, and define $\Delta(y_i^*,y_i)$,

\begin{equation}
\label{equ:loss_multiclass}
\begin{aligned}
\Delta(\bfy_i^*,\bfy_i)=
\left\{\begin{matrix}
1, &if~\bfy_i^*=\bfy_i, \\
0, &otherwise.
\end{matrix}\right.
\end{aligned}
\end{equation}

\item The second data set is SUN data set \cite{Xiao20103485}. The outputs of this data set are the nodes of a tree structure. In this data set, there a 2,000 images, and they belongs to 15 different classes of scenes. The classes are organized as a scene tree. The root node is scene, and it has three child nodes, which are indoor, outdoor land space, and outdoor man-made. These three child nodes have further 15 leaf nodes, which are the 15 classes. Thus there are 19 nodes in the tree in total. The scene tree is shown in figure \ref{fig:tree}. Each image belongs to one of the classes. To represent the image, we extract the HOG features from the image and use them as visual features. In this case, the structured output is a node of the tree. We present a output of the $i$-th data point by using a 19-dimensional binary vector $\bfy_i\in \{1,0\}^{19}$. The $k$-th element of $\bfy_i$ is defined as

\begin{equation}
\begin{aligned}
y_{ik}=
\left\{\begin{matrix}
1, &if~the~k-th~node~is~the~class~of~\bfx_i,\\
&~or~it~is~a~ancestor~of~the~class~of~\bfx_i, \\
0, &otherwise.
\end{matrix}\right.
\end{aligned}
\end{equation}
We also define the joint input-output representation function as $\Phi(\bfx,\bfy) = \bfx\otimes \bfy$. The structured loss function
$\Delta(\bfy_i^*,\bfy_i)$ is defined as the height of the first common ancestor of the predicted output $\bfy_i^*$ and true output $\bfy_i$.

\begin{figure}
  \centering
  \includegraphics[width=0.7\textwidth]{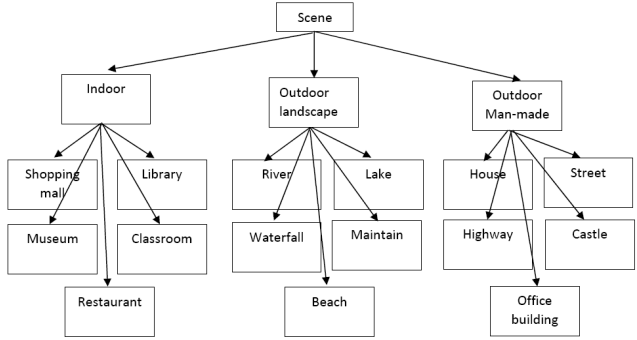}\\
  \caption{Tree structured outputs of SUN data set.}
  \label{fig:tree}
\end{figure}

\item The third data set is a subset of Biocreative data set, provided by the special session of CoNLL2002  \cite{tsochantaridis2005large}. The outputs of this data set is label sequences. This set contains 500 sentences from biomedical papers. Each word in a sentences can be labeled as one of the nine named entities. The problem is to assign a sequence of named entity labels to a sentence. Thus the output of a sentence of $m$ words, $\bfx_i$, is a sequence of labels, $y_i=(y_{i1},\cdots,y_{im})$, where $y_{ik}$ is the label of the $k$-th word. The joint input-output representation function, $\Phi(\bfx_i,y_i)$ is defined as the histogram of state transition and a set of features describing the emissions \cite{tsochantaridis2004support}. The structured loss function to compare a predicted label sequence $y_i^*$ against the truce label sequence $y_i$ is defined as follows,

\begin{equation}
\begin{aligned}
\Delta(y_i^*,y_i)=
\left\{\begin{matrix}
1, &if~y_i^*=y_i, \\
0, &otherwise.
\end{matrix}\right.
\end{aligned}
\end{equation}

\end{itemize}

\subsection{Experiment setup}

To perform the experiment, we employ the 10-fold cross validation. A entire data set is split into ten folds randomly. Each fold is used as a test set in turn. The rest nine folds are combined as a training set. Moreover, we further select two folds from the training set randomly as labeled data set, and leave the rest seven folds as unlabeled data set. The proposed method is applied to the training set to learn the predictive model parameter, and the structured outputs of the unlabeled training data points. Moreover, the learned predictive model are also applied to the test set to predict the structured outputs of the test data points. The prediction performance is evaluated by the average structured loss (ASL) over the test set, $\mathcal{T}$,

\begin{equation}
\begin{aligned}
ASL = \frac{1}{|\mathcal{T}|}\sum_{i:\theta_i\in \mathcal{T}} \Delta(y_i^*,y_i).
\end{aligned}
\end{equation}

\subsection{Experiment results}

In this section, we study the proposed method experimentally. We first compare it to the state-of-the-art semi-supervised structured output prediction methods. Then we study the convergency of the proposed iterative algorithm. Finally, we study how the algorithm performs over different tradeoff parameters.

\subsubsection{Comparison to state-of-the-art}

\begin{figure}
\centering
\subfigure[Core data set]{
\includegraphics[width=0.7\textwidth]{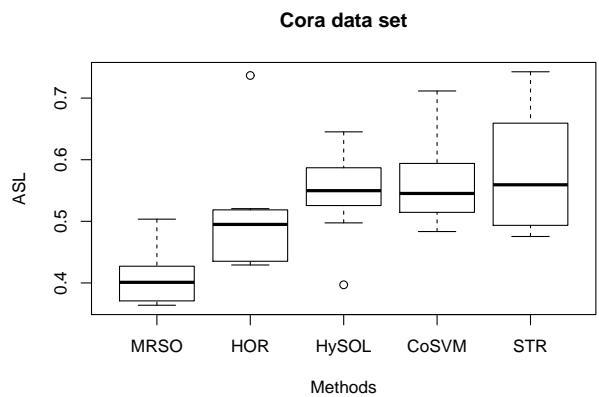}
\label{fig:core}}
\subfigure[SUN data set]{
\includegraphics[width=0.7\textwidth]{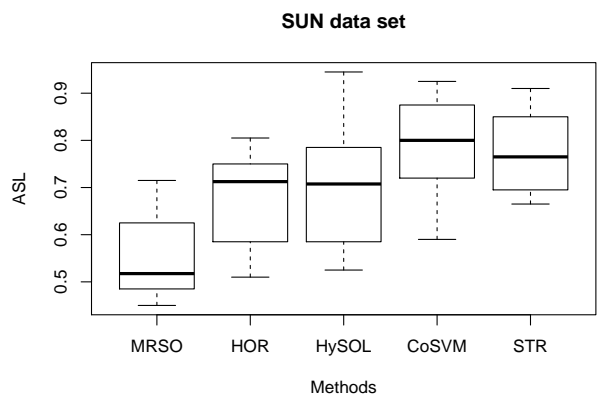}
\label{fig:SUN}}
\subfigure[Biocreative data set]{
\includegraphics[width=0.7\textwidth]{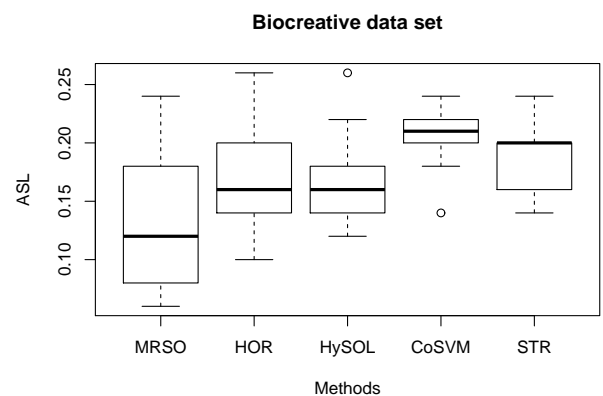}}
\caption{Results of comparison to state-of-the-art.}
\label{fig:comparision}
\end{figure}

We compare the proposed MRSO algorithm against several state-of-the-art semi-supervised learning methods for structured output prediction. We list them as follows:

\begin{itemize}
\item Semi-supervised structured (STR) max-margin optimization method \cite{Altun200533},
\item Co-support vector learning for structured output variables (CoSVM) \cite{Brefeld2006145a},
\item Semi-supervised structured output learning based on a hybrid generative and discriminative models (HySOL) \cite{Suzuki2007791}, and
\item High order regularization for semi-supervised learning of structured output problems (HOR) \cite{Li20143205}.
\end{itemize}
The boxplots of the 10-fold cross validation are given in figure \ref{fig:comparision}. From results in figure \ref{fig:comparision}, we can easily determine that the proposed MRSO algorithm outperforms the other algorithms over all three data sets. For example, in figure \ref{fig:core}, we can see that the median value of the ASL values of the MRSO is as low as about 0.4, while the median ASL of the second best method, HOR, is as high as 0.5. For all other three methods, the media values of ASL are higher than 0.5, which are around 0.55. The outperforming of the proposed algorithm MRSO over the compared methods is even more obvious in \ref{fig:SUN}.  In this figure, only the proposed MRSO method achieves a median value of ASL lower than 0.6, and those median values of the compared methods are higher than 0.7. Moreover, it seems that HOR and HySOL performs better than CoSVM and STR.

\subsubsection{Algorithm convergency}

\begin{figure}[!htb]
  \centering
  \includegraphics[width=0.7\textwidth]{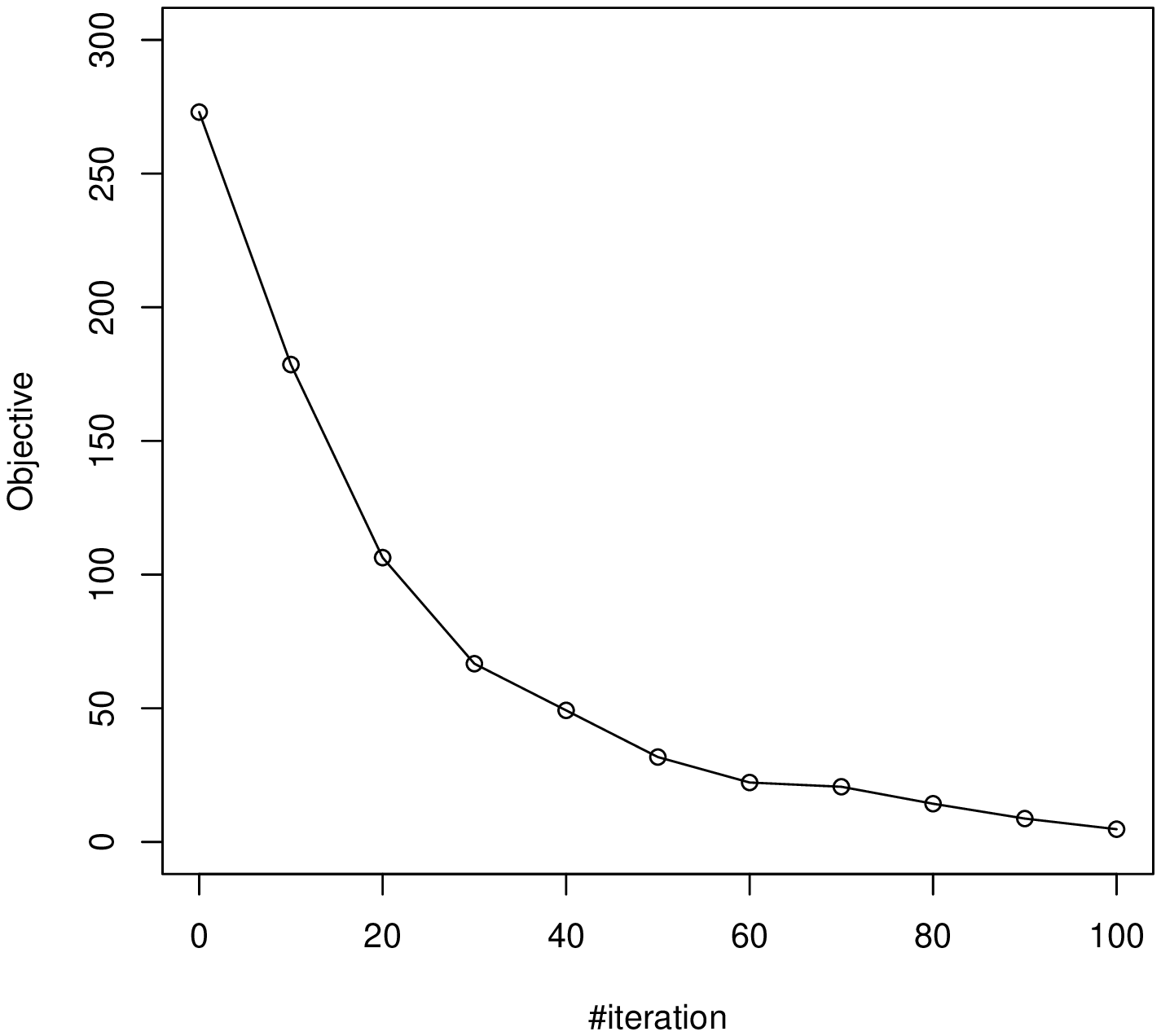}\\
  \caption{Responses of objective function of different iterations.}
  \label{fig:conv}
\end{figure}

The proposed algorithm is an iterative algorithm. We also study the convergency of the algorithm by plotting the responses of the objective function of different iterations. This experiment is conducted over the  Cora data set.
The curve is given in figure \ref{fig:conv}. From this figure, we can observe the iterative algorithm can converge at some point of iteration. For example, the objective decreases significantly from the first iteration to the 60-th iteration, and then the objective stays stable after the 60-the iteration. This indicates the algorithm converges.

\subsubsection{Tradeoff parameter analysis}

\begin{figure}
\centering
\subfigure[$C_1$]{
\label{fig:C1}
\includegraphics[width=0.7\textwidth]{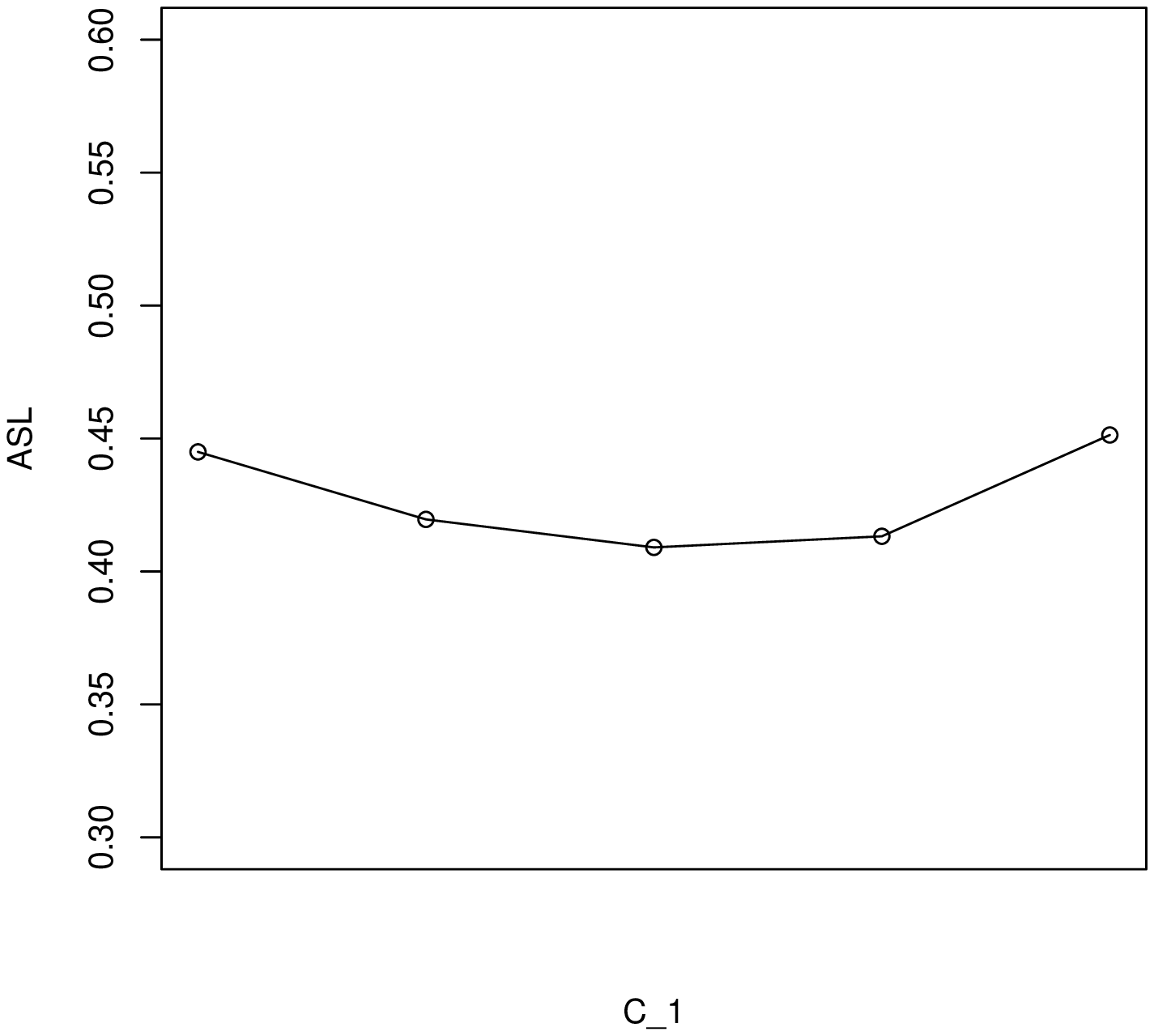}
}\\
\subfigure[$C_2$]{
\label{fig:C2}
\includegraphics[width=0.7\textwidth]{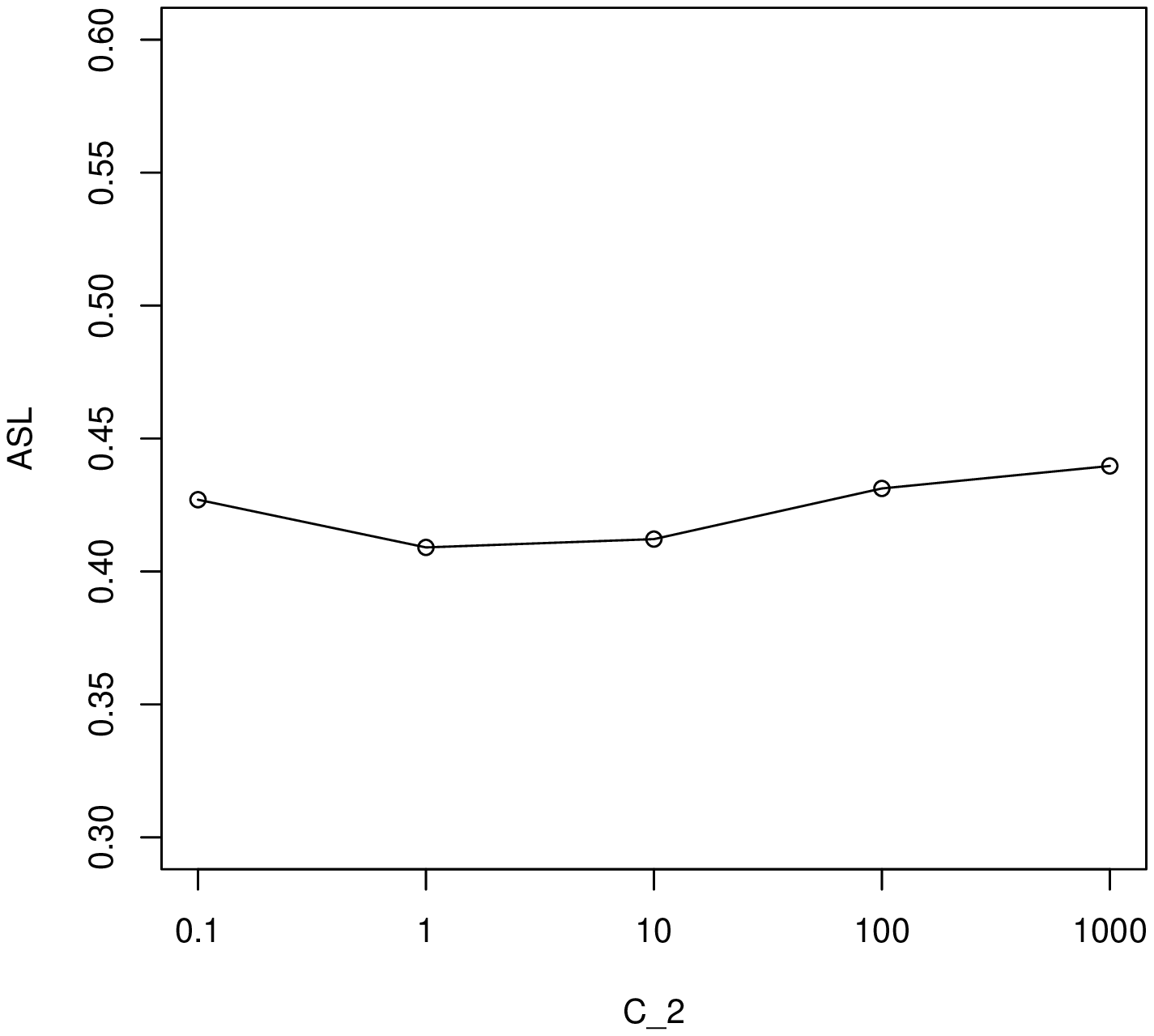}
}\\
\caption{Sensitivity curve of tradeoff parameters.}
\label{fig:param}
\end{figure}

In the objective function of our formulation (\ref{equ:object}), there are two tradeoff parameters, $C_1$ and $C_2$. We also want to know how these parameters effect the performance of our algorithm. To this end, we plot the curve of the different values of ASL of different values of $C_1$  and $C_2$. The curves are shown in figure \ref{fig:param}. Please note that the data in figure \ref{fig:param} is obtained by conducting experiments in Cora data set. From this figure, we can observe that our algorithm is table to both the parameters. In figure \ref{fig:C1}, when the parameter $C_1$ varies from 0.1 to 1000, the range of ALS of MRSO is $[0.40, 0.45]$, and the variance is very small. Moreover, in figure \ref{fig:C2}, we can also observe that the range of ALS of MRSO is $[0.40, 0.43]$ when $C_2$ is varied.

\section{Conclusion and future works}
\label{sec:conclusion}

This paper investigate the problem of semi-supervised structured output prediction. We propose to use the manifold structure to regularize the structured outputs directly. However, in this problem, many training data points only have input feature vectors, while the structured outputs are missing. To solve this problem, we propose a slack structured output for each training data point, either labeled or unlabeled. Moreover, we construct a nearest neighbor graph in the input space to present the manifold structure, and use it to regularize the learning of the slack structured outputs. We impose the slack structured outputs to be consistent to both the manifold structure and the prediction results of a structured output predictor. More specifically, we use a structured loss function to measure how a pair of structured output fits to the manifold distribution. A unified objective is constructed for the learning of both slack structured outputs and the predictive model parameter, and an iterative algorithm is proposed to minimize this objective function. The experiment results show that the proposed algorithm outperforms the state-of-the-art semi-supervised structured output prediction methods.


\begin{thebibliography}{10}
\providecommand{\url}[1]{{#1}}
\providecommand{\urlprefix}{URL }
\expandafter\ifx\csname urlstyle\endcsname\relax
  \providecommand{\doi}[1]{DOI~\discretionary{}{}{}#1}\else
  \providecommand{\doi}{DOI~\discretionary{}{}{}\begingroup
  \urlstyle{rm}\Url}\fi

\bibitem{Abdelouadoud20159}
Abdelouadoud, S., Girard, R., Neirac, F., Guiot, T.: Optimal power flow of a
  distribution system based on increasingly tight cutting planes added to a
  second order cone relaxation.
\newblock International Journal of Electrical Power and Energy Systems
  \textbf{69}, 9--17 (2015)

\bibitem{Altun200533}
Altun, Y., McAllester, D., Belkin, M.: Maximum margin semi-supervised learning
  for structured variables.
\newblock In: Advances in Neural Information Processing Systems, pp. 33--40
  (2005)

\bibitem{Astikainen2011367}
Astikainen, K., Holm, L., Pitk{\"a}nen, E., Szedmak, S., Rousu, J.: Structured
  output prediction of novel enzyme function with reaction kernels.
\newblock Communications in Computer and Information Science \textbf{127 CCIS},
  367--379 (2011)

\bibitem{Braida20154733}
Braida, F., Mello, C.E., Pasinato, M.B., Zimbrao, G.: Transforming
  collaborative filtering into supervised learning.
\newblock Expert Systems with Applications \textbf{42}(10), 4733--4742 (2015)

\bibitem{Brefeld2006145a}
Brefeld, U., Scheffer, T.: Semi-supervised learning for structured output
  variables.
\newblock In: ICML 2006 - Proceedings of the 23rd International Conference on
  Machine Learning, vol. 2006, pp. 145--152 (2006)

\bibitem{Chouman201599}
Chouman, M., Crainic, T.: Cutting-plane matheuristic for service network design
  with design-balanced requirements.
\newblock Transportation Science \textbf{49}(1), 99--113 (2015)

\bibitem{Eronen2015641}
Eronen, V.P., M{\"a}kel{\"a}, M., Westerlund, T.: Extended cutting plane method
  for a class of nonsmooth nonconvex minlp problems.
\newblock Optimization \textbf{64}(3), 641--661 (2015)

\bibitem{Fang2015212}
Fang, Y., Chu, F., Mammar, S., Shi, Q.: A new cut-and-solve and cutting plane
  combined approach for the capacitated lane reservation problem.
\newblock Computers and Industrial Engineering \textbf{80}, 212--221 (2015)

\bibitem{Feng2015}
Feng, J., Wang, J., Zhang, H., Han, Z.: Fault diagnosis method of joint fisher
  discriminant analysis based on the local and global manifold learning and its
  kernel version.
\newblock IEEE Transactions on Automation Science and Engineering  (2015).
\newblock \doi{10.1109/TASE.2015.2417882}

\bibitem{Han20141665}
Han, Y., Wei, X., Cao, X., Yang, Y., Zhou, X.: Augmenting image descriptions
  using structured prediction output.
\newblock IEEE Transactions on Multimedia \textbf{16}(6), 1665--1676 (2014)

\bibitem{Ho2015}
Ho, S., Dai, P., Rudzicz, F.: Manifold learning for multivariate
  variable-length sequences with an application to similarity search.
\newblock IEEE Transactions on Neural Networks and Learning Systems  (2015).
\newblock \doi{10.1109/TNNLS.2015.2399102}

\bibitem{Joachims20061}
Joachims, T.: Structured output prediction with support vector machines.
\newblock Lecture Notes in Computer Science (including subseries Lecture Notes
  in Artificial Intelligence and Lecture Notes in Bioinformatics) \textbf{4109
  LNCS}, 1--7 (2006)

\bibitem{Kajdanowicz2011333}
Kajdanowicz, T., Wozniak, M., Kazienko, P.: Multiple classifier method for
  structured output prediction based on error correcting output codes.
\newblock Lecture Notes in Computer Science (including subseries Lecture Notes
  in Artificial Intelligence and Lecture Notes in Bioinformatics) \textbf{6592
  LNAI}(PART 2), 333--342 (2011)

\bibitem{Kim2010649}
Kim, M., Pavlovic, V.: Structured output ordinal regression for dynamic facial
  emotion intensity prediction.
\newblock Lecture Notes in Computer Science (including subseries Lecture Notes
  in Artificial Intelligence and Lecture Notes in Bioinformatics) \textbf{6313
  LNCS}(PART 3), 649--662 (2010)

\bibitem{Li20143205}
Li, Y., Zemel, R.: High order regularization for semi-supervised learning of
  structured output problems.
\newblock In: 31st International Conference on Machine Learning, ICML 2014,
  vol.~4, pp. 3205--3217 (2014)

\bibitem{Lorente201517}
Lorente, D., Escandell-Montero, P., Cubero, S., G{\'o}mez-Sanchis, J., Blasco,
  J.: Visible-nir reflectance spectroscopy and manifold learning methods
  applied to the detection of fungal infections on citrus fruit.
\newblock Journal of Food Engineering \textbf{163}, 17--24 (2015)

\bibitem{luo2011based}
Luo, J., Brodsky, A.: An em-based multi-step piecewise surface regression
  learning algorithm.
\newblock In: The seventh international conference on data mining (WORLDCOMP
  DMIN 11), pp. 286--292 (2011)

\bibitem{Oonk201580}
Oonk, S., Spijker, J.: A supervised machine-learning approach towards
  geochemical predictive modelling in archaeology.
\newblock Journal of Archaeological Science \textbf{59}, 80--88 (2015)

\bibitem{senaimag08}
Sen, P., Namata, G.M., Bilgic, M., Getoor, L., Gallagher, B., Eliassi-Rad, T.:
  Collective classification in network data.
\newblock AI Magazine \textbf{29}(3), 93--106 (2008)

\bibitem{Su201038}
Su, H., Heinonen, M., Rousu, J.: Structured output prediction of anti-cancer
  drug activity.
\newblock Lecture Notes in Computer Science (including subseries Lecture Notes
  in Artificial Intelligence and Lecture Notes in Bioinformatics) \textbf{6282
  LNBI}, 38--49 (2010)

\bibitem{Suzuki2007791}
Suzuki, J., Fujino, A., Isozaki, H.: Semi-supervised structured output learning
  based on a hybrid generative and discriminative approach.
\newblock In: EMNLP-CoNLL 2007 - Proceedings of the 2007 Joint Conference on
  Empirical Methods in Natural Language Processing and Computational Natural
  Language Learning, pp. 791--800 (2007)

\bibitem{tsochantaridis2004support}
Tsochantaridis, I., Hofmann, T., Joachims, T., Altun, Y.: Support vector
  machine learning for interdependent and structured output spaces.
\newblock In: Proceedings of the twenty-first international conference on
  Machine learning, p. 104. ACM (2004)

\bibitem{tsochantaridis2005large}
Tsochantaridis, I., Joachims, T., Hofmann, T., Altun, Y.: Large margin methods
  for structured and interdependent output variables.
\newblock In: Journal of Machine Learning Research, pp. 1453--1484 (2005)

\bibitem{wang2014effective}
Wang, H., Wang, J.: An effective image representation method using kernel
  classification.
\newblock In: Tools with Artificial Intelligence (ICTAI), 2014 IEEE 26th
  International Conference on, pp. 853--858. IEEE (2014)

\bibitem{wang2010data}
Wang, J., Wan, J., Liu, Z., Wang, P.: Data mining of mass storage based on
  cloud computing.
\newblock In: Grid and Cooperative Computing (GCC), 2010 9th International
  Conference on, pp. 426--431. IEEE (2010)

\bibitem{wang2015multiple}
Wang, J., Wang, H., Zhou, Y., McDonald, N.: Multiple kernel multivariate
  performance learning using cutting plane algorithm.
\newblock In: Systems, Man and Cybernetics (SMC), 2015 IEEE International
  Conference on. IEEE (2015)

\bibitem{wang2015representing}
Wang, J., Zhou, Y., Yin, M., Chen, S., Edwards, B.: Representing data by sparse
  combination of contextual data points for classification.
\newblock In: Advances in Neural Networks--ISNN 2015. Springer (2015)

\bibitem{wang2014next}
Wang, K., Zhou, X., Chen, H., Lang, M., Raicu, I.: Next generation job
  management systems for extreme-scale ensemble computing.
\newblock In: Proceedings of the 23rd international symposium on
  High-performance parallel and distributed computing, pp. 111--114. ACM (2014)

\bibitem{wang2014optimizing}
Wang, K., Zhou, X., Li, T., Zhao, D., Lang, M., Raicu, I.: Optimizing load
  balancing and data-locality with data-aware scheduling.
\newblock In: Big Data (Big Data), 2014 IEEE International Conference on, pp.
  119--128. IEEE (2014)

\bibitem{wang2015towards}
Wang, K., Zhou, X., Qiao, K., Lang, M., McClelland, B., Raicu, I.: Towards
  scalable distributed workload manager with monitoring-based weakly consistent
  resource stealing.
\newblock In: Proceedings of the 24rd international symposium on
  High-performance parallel and distributed computing, pp. 219--222. ACM (2015)

\bibitem{wang2012mathematical}
Wang, Y., Yang, T., Ma, Y., Halade, G.V., Zhang, J., Lindsey, M.L., Jin, Y.F.:
  Mathematical modeling and stability analysis of macrophage activation in left
  ventricular remodeling post-myocardial infarction.
\newblock BMC genomics \textbf{13}(Suppl 6), S21 (2012)

\bibitem{Wu20092087}
Wu, Y., Yuan, Z., Liu, Y., Zheng, N.: Discriminative structured outputs
  prediction model and its efficient online learning algorithm.
\newblock In: 2009 IEEE 12th International Conference on Computer Vision
  Workshops, ICCV Workshops 2009, pp. 2087--2094 (2009)

\bibitem{xia2015study}
Xia, P., Liu, B., Sun, Y., Chen, C.: Reciprocal recommendation system for
  online dating.
\newblock In: Proceedings of the 2015 IEEE/ACM International Conference on
  Advances in Social Networks Analysis and Mining. ACM (2015)

\bibitem{Xiao20103485}
Xiao, J., Hays, J., Ehinger, K., Oliva, A., Torralba, A.: Sun database:
  Large-scale scene recognition from abbey to zoo.
\newblock In: Proceedings of the IEEE Computer Society Conference on Computer
  Vision and Pattern Recognition, pp. 3485--3492 (2010).
\newblock \doi{10.1109/CVPR.2010.5539970}

\bibitem{Xing2015395}
Xing, X., Wang, K., Lv, Z., Zhou, Y., Du, S.: Fusion of local manifold learning
  methods.
\newblock IEEE Signal Processing Letters \textbf{22}(4), 395--399 (2015)

\bibitem{cross13}
Xu, L., Zhan, Z., Xu, S., Ye, K.: Cross-layer detection of malicious websites.
\newblock In: Proceedings of the third ACM conference on Data and application
  security and privacy, pp. 141--152. ACM (2013)

\bibitem{web14}
Xu, L., Zhan, Z., Xu, S., Ye, K.: An evasion and {Counter-Evasion} study in
  malicious websites detection.
\newblock In: 2014 IEEE Conference on Communications and Network Security (CNS)
  (IEEE CNS 2014). San Francisco, USA (2014)

\bibitem{zhang2012automated}
Zhang, H., Jiao, Y., Zhang, Y., Shimada, K.: Automated segmentation of cerebral
  aneurysms based on conditional random field and gentle adaboost.
\newblock Mesh Processing in Medical Image Analysis 2012 pp. 59--69 (2012)

\bibitem{zhang2011empirical}
Zhang, S., Caragea, D., Ou, X.: An empirical study on using the national
  vulnerability database to predict software vulnerabilities.
\newblock In: Database and Expert Systems Applications, pp. 217--231. Springer
  Berlin Heidelberg (2011)

\end{thebibliography}

\end{document}